\documentclass[10pt,twocolumn,letterpaper]{article}

\usepackage{cvpr}
\usepackage{times}
\usepackage{epsfig}
\usepackage{graphicx}
\usepackage{amsmath}
\usepackage{amssymb}

\usepackage{multirow}
\usepackage{makecell}
\usepackage{booktabs}

\usepackage[pagebackref=true,breaklinks=true,letterpaper=true,colorlinks,bookmarks=false]{hyperref}

\cvprfinalcopy  

\def\cvprPaperID{9123} 

\ifcvprfinal\fi
\begin{document}
	
	\title{First image then video: A two-stage network for spatiotemporal video denoising}
	
	\author{Ce Wang, S. Kevin Zhou, Zhiwei Chen \\
		Nankai University\\
		Center for Combinatorics, Nankai University, Tianjin 300071, P.R. China\\
		{1120150001@mail.nankai.edu.cn}}
	\author{Ce Wang\\
	Z$^2$Sky Technologies Inc. \\ \& NanKai University\\
	{1120150001@mail.nankai.edu.cn}
	\and
	S. Kevin Zhou\\
	Chinese Academy of Sciences(CAS) \\ \& Peng Cheng Laboratory(PCL)\\
	{zhoushaohua@ict.ac.cn}
	\and
    Zhiwei Cheng\\
    Z$^2$Sky Technologies Inc.\\
    {chengzw19@163.com}
	}
	
	\maketitle
	
	\begin{abstract}
		%
		%

		Video denoising is to remove noise from noise-corrupted data, thus recovering true signals via spatiotemporal processing. Existing approaches for spatiotemporal video denoising tend to 
		suffer from motion blur artifacts, that is, the boundary of a moving object tends to appear 
		blurry especially when the object undergoes a fast motion, causing optical flow calculation to break down. In this paper, we address this challenge by 
		designing a first-image-then-video two-stage denoising neural network, consisting of an image denoising module for spatially reducing intra-frame noise followed by a regular spatiotemporal video denoising module. The 
		intuition is simple yet powerful and effective: the first stage of image denoising effectively reduces the 
		noise level and, therefore, allows the second stage of spatiotemporal denoising for better modeling and learning everywhere, including along the moving object boundaries. This two-stage network, when 
		trained in an end-to-end fashion, yields the state-of-the-art performances on 
		the video denoising benchmark Vimeo90K dataset in terms of both denoising quality and computation. It also enables an unsupervised approach that achieves comparable performance to existing supervised approaches.
		
	\end{abstract}
	
	\begin{figure*}
		\begin{center}
			\includegraphics[width=\textwidth]{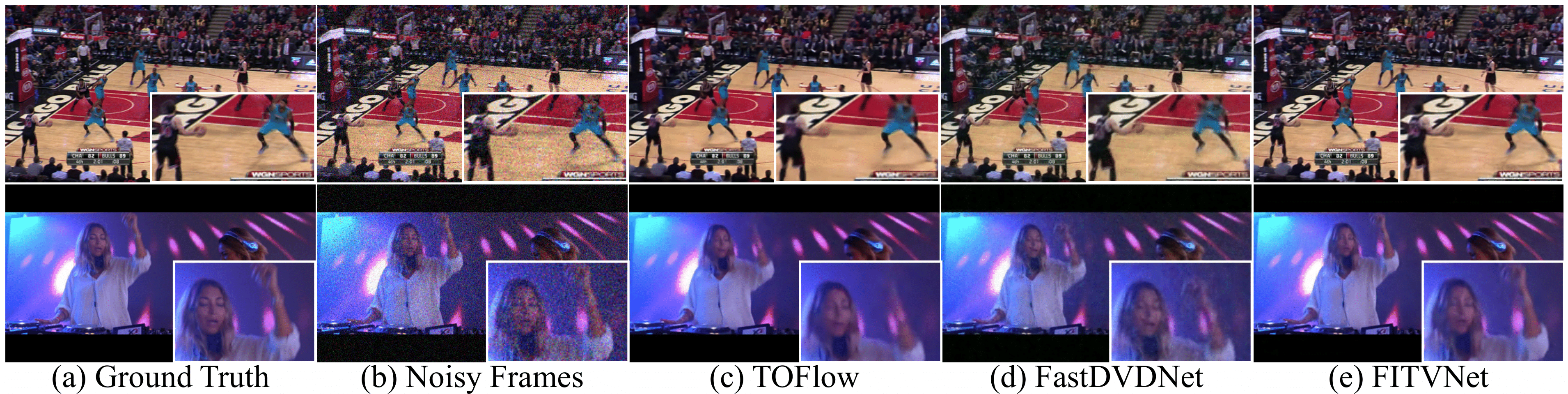}
		\end{center}
		\caption{Samples of denoised frames of TOFlow, FastDVDNet, and our FITVNet model. In these cases, 
			where an object undergoes a fast motion, TOFlow (c) fails to 
			estimate motion accurately enough for effective denoising and noise-reduced frames appear blurry along object boundaries. Although FastDVDNet (d) avoids optical flow calculation, they still could not output better results as in  since their spatiotemporal
			denoising block could hardly handle with both spatial noise and temporal deformation. 
			In contrast, our FITV model (e) shows high performance even on these objects with a prior image denoising module for spatially reducing intra-frame noise.}
		\label{fig-pic-intro}
	\end{figure*}
	\section{\textbf{Introduction}}
	
	Image denoising is an important task to recover clean signal $I$ from noisy signal
	$\hat{I}=I+N$, where $N$ is some kind of noise added pixel by pixel. As an active
	research area for a long time, traditional methods \cite{dabov2007image} \cite{lefkimmiatis2017non}
	have been proposed and achieved state-of-the-art results on many benchmarks. 
	To further deal with real-world images without ground truth, unsupervised methods utilizing CNNs 
	\cite{batson2019noise2self}\cite{krull2019noise2void}\cite{lehtinen2018noise2noise}
	\cite{xu2019noisy} attempt to denoise only with noisy images and successfully train networks to 
	approximate target data distribution. Although these methods always pose more assumptions 
	and requirements on their inputs, they have achieved quite success.
	
	As a more difficult task with the addition of temporal information, video denoising aims to 
	recover clean signals $V = \{{I}_{1}, {I}_{2}, \dots, {I}_{T}\}$ from a noisy video
	$\hat{V} = V + \widetilde{N}$, where $\widetilde{N} = \{{N}_{1},{N}_{2}, \dots, {N}_{T}\}$
	and $\{{N}_{t}\}_{t=1}^{T}$ is the noise of each frame. In addition, video denoising can gain an advantage by proper modeling of inter-frame temporal deformation exists due to object motion in videos. For example, state-of-the-art TOFlow~\cite{xue2019video} approach tries to perform motion analysis and video denoising 
	simultaneously in an end-to-end trainable network. However, we often
	observe blurry object boundaries in their denoised results when these objects undergo 
	a fast motion as shown in Figure~\ref{fig-pic-intro} (c) or when the foreground and background between these boundaries are of weak contrast due to say the low light environment. The blurry phenomenon happens because motion estimation breaks down 
	for these objects in videos even in such a well-designed task-oriented flow estimation method.
	
	Furthermore, the success application of convolutional neural networks (CNNs) to image denoising encourages researchers
	to design suitable neural models for video denoising. Although only invoking
	image denoising algorithm for each frame is not effective, the spatiotemporal denoising
	like V-BM4D\cite{maggioni2012video}, which extends BM3D~\cite{dabov2007image} by searching
	similar patches in both spatial and temporal dimensions, is instructive. 
	FastDVDNet\cite{tassano2019fastdvdnet}, one state-of-the-art video denoising algorithm, 
	extends flow-based DVDNet~\cite{tassano2019dvdnet} by employing modified 
	U-Net~\cite{ronneberger2015u} architectures to realize spatiotemporal 
	denoising with the powerful representation of CNNs. This avoids explicit estimation of objection
	motion, but the results shown in Figure~\ref{fig-pic-intro}(d) demonstrate that the spatiotemporal denoising block still 
	could not handle high-speed moving objects well because spatial noise distribution and temporal deformation are very different and dealing with them together in a mixed fashion is too challenging. 
	The limited capability of spatiotemporal denoising block brings the main 
	motivation of our work: Can we address this challenge by separating CNN-based spatiotemporal denoising into a two-stage procedure, 
	in which the first module tries to spatially reduce the intra-frame noise in ${N}_{t}$ and the
	second module, still as regular spatiotemporal inter-frame video denoising, is able to handle these cases
	with the help of ``processed" inputs of the first stage? 
	
	In order to formulate such a first-image-then-video (FITV) two-stage denoising process in an end-to-end network, 
	we propose our neural model, called FITVNet. We realize the first module with an image-to-image network architecture, which reduces the noise within each single image/frame via spatial processing, and the second module as a regular spatiotemporal 
	video denoising network. These two modules are jointly supervised by a proposed loss function with a balanced learning ratio between each other in different training phases.
	With such a training procedure, output features of the first stage show high denoising
	performance for reducing intra-frame noise, especially along object boundaries where TOFlow and FastDVD fail to recover.
	In addition to designing such an end-to-end network for video denoising, 
	we empirically demonstrate the effectiveness of such a two-stage denoising method via extensive
	comparison with state-of-the-art video denoising approaches in terms of both denoising quality and computation. We also introduce an unsupervised video denoising approach. 
	
	The main contributions of our work are as below. 
	\begin{itemize}
		\item We propose a two-stage video denoising algorithm which reduces intra-frame noise spatially first and then
		applies spatiotemporal video denoising, which achieves high denoising performance everywhere including boundaries of fast-moving objects.
		\item We integrate these two stages in an end-to-end trainable network with one supervision signal for these two different modules. We also explore incorporating different types of supervision signals.
		\item We achieve state-of-the-art video denoising results on the benchmarking Vimeo90K dataset in terms of both denoising quality and computation. Our completely unsupervised video denoising approach achieves comparable performance to current supervised approaches.
	\end{itemize}

	\begin{figure}[t]
		\begin{center}
			\includegraphics[width=\columnwidth]{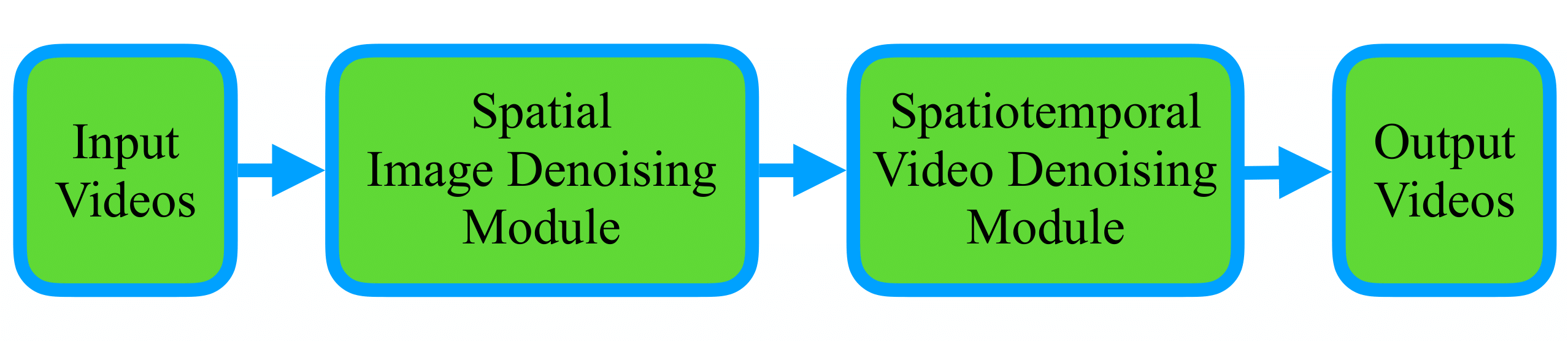}
		\end{center}
		\caption{The FTIV denoising framework of our model. Its simple idea of adding a single-image based denoising module before regular video denoising is proven effective. }
		\label{fig-framework}
	\end{figure}
	\section{Previous Works}
	
	\subsection{Image denoising}
	Image denoising has been explored for a long time and recent methods
	~\cite{burger2012image}~\cite{lefkimmiatis2017non}~\cite{mao2016image}~\cite{tai2017memnet} 
	~\cite{xie2012image}~\cite{zhang2017beyond}~\cite{lefkimmiatis2018universal}
	~\cite{weigert2017isotropic} make use of the representative capability of CNNs
	rather than traditional methods~\cite{dabov2007image}~\cite{roth2005fields}~\cite{tappen2007learning}. 
	These methods take pairs of noisy and clean images as inputs
	to train their models. Then models could learn a mapping function from noisy image 
	distribution ${P}_{Y} = {P}_{X} + {P}_{noise}$ to real image distribution ${P}_{X}$ via
	minimizing a loss function between their outputs and corresponding clean images, 
	and achieve better results than traditional state-of-the-art models such as BM3D~\cite{dabov2007image}.
	
	But clean images are expensive to obtain when facing with real-world scenarios.
	To tackle such a difficulty, Noise2Noise~\cite{lehtinen2018noise2noise} utilizes the relationship between the ${P}_{Y}$ and ${P}_{X}$ distributions and trains its model using pairs of both noisy images of the same scene. Following the similar vein, Noise2Void~\cite{krull2019noise2void} Noise2Self~\cite{batson2019noise2self}, and Noise-As-Clean~\cite{xu2019noisy} 
	have recently been proposed, which alleviates the hard requirement used in Noise2Noise that pair images should be with the same scene and independently sampled with the same noise distribution. In this work, we will incorporate the idea of Noise2Noise into the FITVNet.
	
	\subsection{Video denoising}
	Although image denoising has attracted so much research interest, video denoising
	is still under-explored. Directly applying image denoising algorithm to each video frame fails to capture the temporal relationship between consecutive frames. To handle with the key problem of motion estimation in video denoising, traditional methods~\cite{revaud2015epicflow}~\cite{xu2017accurate} are used for 
	flow calculation in video denoising. Later works~\cite{dosovitskiy2015flownet}~\cite{jason2016back}~\cite{ranjan2017optical}
	propose end-to-end networks for flow estimation. Then TOFlow~\cite{xue2019video} and DVDNet~\cite{tassano2019dvdnet} further simultaneously predict optical flows and denoise frames 
	instead of separating these two tasks as in~\cite{niklaus2018context}, 
	and obtain state-of-the-art results that are better 
	than V-BM4D~\cite{maggioni2012video}, which is similar to VNLB~\cite{arias2018video} 
	as extensions of BM3D~\cite{dabov2007image}. 
	However, image warping between neighbouring 
	frames using the estimated flow field is time consuming. Even worse, the flow calculation tends to be inaccurate when an object undergoes fast motion or when the image contrast between the foreground and background is weak, leading to the over-smoothness and loss of details along object boundaries in the denoised frames.
	Alternatively, ViDeNN~\cite{claus2019videnn} and 
	FastDVDNet~\cite{tassano2019fastdvdnet} propose one end-to-end network
	to realize spatiotemporal denoising as \cite{huang2016temporally}
	~\cite{newson2014video}~\cite{wexler2004space} have done in video inpainting.	
	Both of them feed pairs of noisy and clean videos to networks to successively 
	reduce noise in input frames. Similarly, NLNet~\cite{davy2018non} fuses a CNN 
	with a self-similarity search strategy in video denoising and VINet~\cite{kim2019deep} successfully 
	applies CNNs in video inpainting. 
	Furthermore, \cite{ehret2019model} proposes a blind denoising approach based on 
	DnCNN~\cite{zhang2017beyond} and Noise2Noise, which motivates us 
	to utilize unsupervised image denoising algorithm in video denoising process. In this work, we will go beyond one spatiotemporal video denoising network and propose a two-stage FITVNet for both supervised and unsupervised video noising. 
	
	\begin{figure*}
		\begin{center}
			\includegraphics[width=\textwidth]{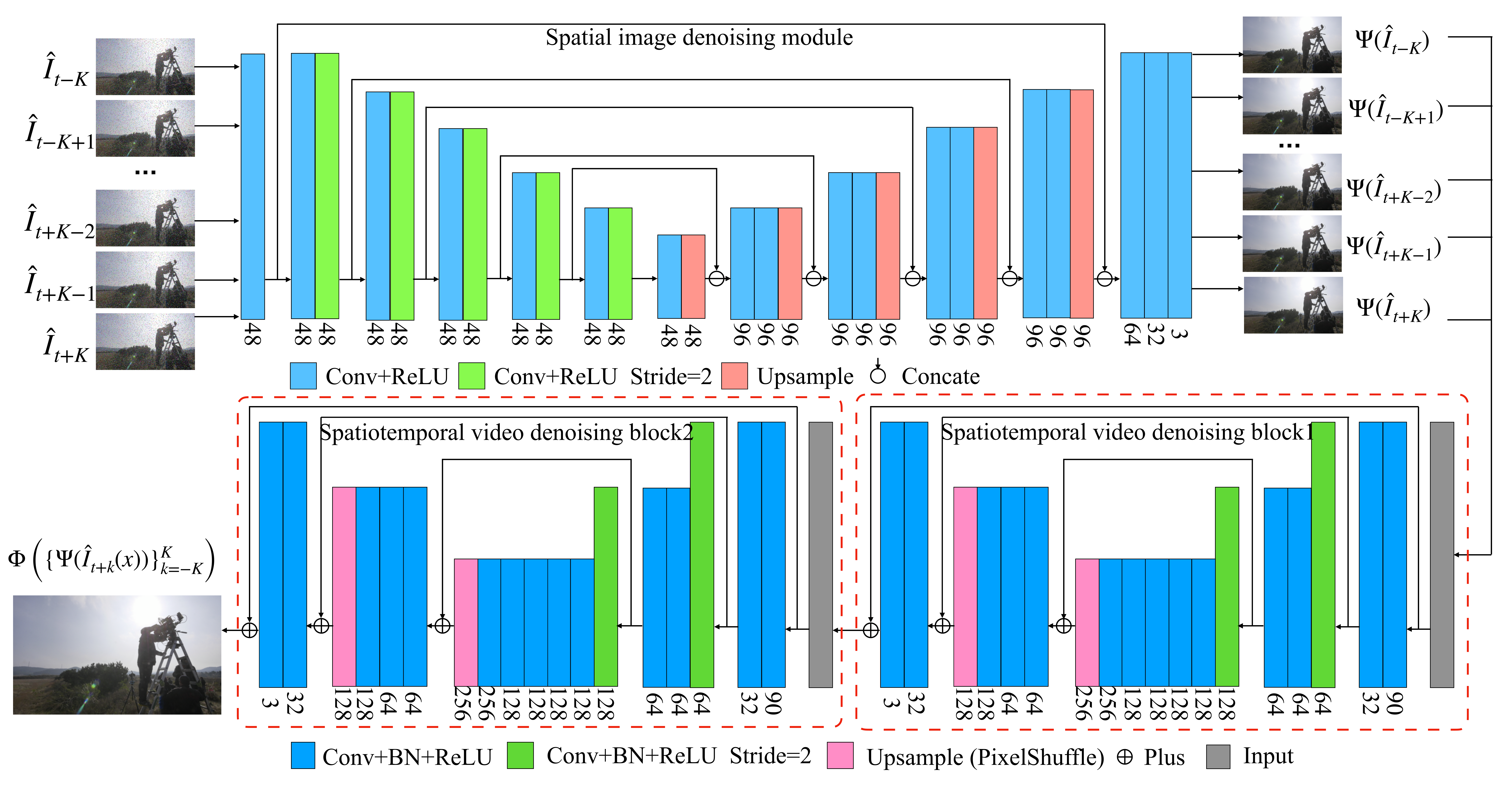}
		\end{center}
		\caption{The whole flow chart of our model. In the first stage, we remove intra-frame
			noise spatially via image denoising module. Then we apply spatiotemporal denoising module
			in the second stage to remove the rest of noise via spatiotemporal processing.}
		\label{fig-model}
	\end{figure*}
	
	\section{Method}
	To describe our model clearly, we first formulate the theoretical background of FITVNet in this section. 
	Then we show details about these two denoising stages, including network architectures, 
	loss function employed for these two parts, different variants of our FITVNet, etc. 
	Figure~\ref{fig-framework} visualizes the diagram of FITVNet.
	\subsection{Theoretical background}
	As mentioned before, object motion is an essential information between neighbouring frames in video 
	denoising and optical flow is always employed to estimate the motion. Consider $2K$ neighbouring frames 
	${I}_{t-K}(x), {I}_{t-K+1}(x), \dots, {I}_{t+K-1}(x), {I}_{t+K}(x)$ around the reference frame 
	${I}_{t}(x)$ in a clean video, the relationship between each neighbouring frame and the reference frame 
	is formulated as follows, assuming ``constancy of brightness":
	\begin{equation}
	{I}_{t+k}(x) = {I}_{t}(x+{\delta}_{k}x);  \quad k \in \{-K,\dots,K-1, K\}, 
	\end{equation}
	where ${\delta}_{k}x$ denotes the estimated optical flow field between these two frames.
	When the optical flow field is of small magnitude, we use the first-order Taylor approximation so that temporal deformation becomes ``additive":
	\begin{equation}
	{I}_{t+k}(x) \approx {I}_{t}(x)+J_t(x){\delta}_{k}x, \end{equation}
	where $J_t(x)$ is the Jacobian matrix. In real-world, we observe noisy frames: 
	\begin{equation}
	\begin{split}
	{\hat{I}}_{t+k}(x) &= {I}_{t+k}(x) + {N}_{k}(x) \\
	&= {I}_{t}(x+{\delta}_{k}x) + {N}_{k}(x)\\
	&\approx {I}_{t}(x)+J_t(x){\delta}_{k}x+ {N}_{k}(x),\ 
	\end{split}
	\end{equation}
	where ${N}_{k}(x)$ is noise added for each frame, such as additive white Gaussian noise (AWGN).
	
	What DVDNet~\cite{tassano2019dvdnet}  achieves in this theoretical framework is to learn an 
	approximate mapping function $T$. It essentially incorporates two functions: the first is to estimate optical 
	flow ${\delta}_{k}x$ between observed neighbouring frames and reference frame, and the second is to  
	utilize these intermediate variables to predict clean reference 
	frame ${I}_{t}$, which can be represented in the following formula: 
	\begin{equation}
	{I}_{t}(x) = T\left(\{{\hat{I}}_{t+k}(x-{\delta}_{k}x)\}_{k=-K}^{K}\right).
	\end{equation}
	While both functions are implemented using two isolated networks, TOFlow~\cite{xue2019video} takes a further step that achieves simultaneous  estimation of optical 
	flow and fusion of warped frames via end-to-end learning, thus obtaining state-of-the-art performance. However, Figure~\ref{fig-pic-intro} shows that precise flow estimation, even for TOFlow, is hard. 
	
	An alternative method, such as FastDVDNet, applies spatiotemporal
	denoising, skipping these sub-optimal flow calculation. It performs the following formula: 
	\begin{equation}
	{I}_{t}(x) = \Phi\left(\{{\hat{I}}_{t+k}(x)\}_{t=-K}^{K}\right), 
	\end{equation}
	where $\Phi$ performs spatiotemporal processing and is approximated by CNNs. However, the failure in denoising fast-moving objects with apparent,  
	blurry object boundaries, motivates us to sort the spatiotemporal denoising into
	two-stage denoising process as introduced before. More formally, what our FITVNet really does is to
	remove noise in ${\hat{I}}_{t+k}$ in advance via a function $\Psi$, followed by $\Phi$ :
	\begin{equation}
	\begin{split}
	{I}_{t}(x) =\Phi\left( \{\Psi ( {\hat{I}}_{t+k}(x) ) \}_{t=-K}^{K} \right).
	\end{split}
	\end{equation}
	
	With our prior image denoising results as inputs, denoted by $\{\Psi ( {\hat{I}}_{t+k}(x) ) \}_{t=-K}^{K}$, 
	regular spatiotemporal denoising function $\Phi$ deals with ``cleaner" frames, which makes the task less challenging.

	\subsection{The spatial image denoising stage}
	For the first image denoising stage, we use a modified architecture 
	of Noise2Noise~\cite{lehtinen2018noise2noise} to approximate the
	above mapping function $\Psi$, which tries to spatially remove intra-frame  noise in an input 
	sequence of $2K+1=5$ frames. The module is composed of $2K+1$ networks that share the same architectures and parameters as shown in Figure~\ref{fig-model}. 
	Each network, which processes its corresponding frame, is trained with
	the following ${l}_{2}$ loss function: 
	\begin{equation}\label{loss-sd}
	{\cal L}_{pd} = \Sigma_{k=-K}^{K} \| {\Psi}_{\theta}({\hat{I}}_{t+k}) - {I}_{t+k} \|_2, 
	\end{equation}
	where $\{{\hat{I}}_{t+k}\}_{k=-K}^{K}$ are input frames, $\{{\Psi}_{\theta}\}_{k=-K}^{K}$ 
	are mapping functions of the corresponding frames, ${\theta}$ is network parameters of this module, 
	and $\{{I}_{t+k}\}_{k=-K}^{K}$ denote the target supervision signal of input frames.
	
	Especially for the network architecture of this module, we replace original pooling layers 
	with convolutional layers with a stride of two in order to capture more contexts. This helps reduce noise level, especially for boundaries of 
	moving objects so that the following spatiotemporal module learns better 
	everywhere with these pre-denoised frames.
	
	To test the performance of our model with different target signals, we 
	have implemented the following three models with different loss functions ${\cal L}_{pd}$: 
	\begin{itemize}
		\item \underline{FITVNet (base)}: without supervision signal of ${L}_{pd}$ introduced before;
		\item \underline{FITVNet (+jsn)}: with noisy frames as the target signal in ${\cal L}_{pd}$ just like what
		Noise2Noise~\cite{lehtinen2018noise2noise} has done and ${\cal L}_{pd}$ becomes:
		\begin{equation}\label{loss-sd-jsn}
		{\cal L}_{pd} = \Sigma_{k=-K}^{K} \| {\Psi}_{\theta}({\hat{I}}_{t+k}) - {\tilde{I}}_{t+k} \|_2, 
		\end{equation}
		where ${\tilde{I}}_{t+k}$ is noisy frames with Gaussian noise i.i.d to the noise of ${\hat{I}}_{t+k}$; and
		\item \underline{FITVNet (+jsc)}: with clean frames as the target signal in ${\cal L}_{pd}$. 
	\end{itemize}
	Note that other unsupervised image denoising methods such as Noise2Void~\cite{krull2019noise2void} Noise2Self~\cite{batson2019noise2self}, and Noise-As-Clean~\cite{xu2019noisy} can also be used, which we will explore in future. 
	
	Furthermore, the success of FITVNet (+jsn), which trains our first module in an unsupervised manner, unleashes 
	the potential of our model to a completely unsupervised video denoising scenario as discussed in Section 3.4.
	\subsection{The spatiotemporal video denoising stage}
	
	In spatiotemporal video denoising stage, we use another network to capture temporal information between neighbouring frames, that is, it takes
	more frames as inputs to utilize object motion information among these frames. Concretely, the module takes $2K+1$ consecutive prior denoised frames 
	$\{ \Psi({\hat{I}}_{t-K}), \dots, \Psi({\hat{I}}_{t+K-1}), \Psi({\hat{I}}_{t+K} )\}$, 
	among which the central one is chosen as the reference frame to be denoised, as inputs and 
	clean frame ${I}_{t}$ as ground truth.
	Meanwhile, the ${l}_{2}$ loss is used to supervise
	the learning process of the module as in (\ref{loss-re}):
	\begin{equation}\label{loss-re}
	{\cal L}_{st} = \Sigma_{i} \| {\Phi}_{ \gamma}\left( \{{\Psi}_{\theta} ( {\hat{I}}_{t+k}(x) ) \}_{k=-K}^{K} \right) - {I}_{t} \|_2, 
	\end{equation}
	where ${\Phi}_{\gamma}$ is the mapping function of regular video denoising block, 
	and ${I}_{t}$ denotes the clean reference frame. However, the output features of 
	pre-denoising module would influence performance of the current spatiotemporal 
	denoising module, so we add a decay coefficient $\alpha$ to balance ${\cal L}_{pd}$ with ${\cal L}_{st}$
	in the jointly supervised manner, and the final loss function for our model is as follows: 
	\begin{equation}
	\frac{\alpha}{e} {\cal L}_{pd} + {\cal L}_{st}.
	\end{equation}
	where $e$ denotes the total number of epochs before this iteration.
	
	Unlike flow-based methods, such as PWC-Net~\cite{sun2018pwc} and TOFlow~\cite{xue2019video}, such CNN-based spatiotemporal denoising network needs to model spatial noise and temporal deformation 
	within one module. We realize this module via the method employed in FastDVDNet~\cite{tassano2019fastdvdnet}, which 
	uses one block to process every three consecutive frames in the totally five frames and 
	then feed concatenated output features of the first block to the other block as shown in 
	Figure \ref{fig-model}. This avoids explicit flow calculation which is sub-optimal in
	these flow-based models and the experiments show that the representative capability
	of CNNs help avoid boundary blurry appearing in TOFlow with the help of prior image denoising. Furthermore, the expensive computation of warping operations in flow-based methods is also eliminated.
	
	\subsection{Unsupervised FITVNet} \label{unsupervised_FITVNet}
	With the success of applying prior image denoising stage before regular video denoising stage
	and training the module with noisy images as in FITVNet (+jsn), we have the opportunity to design a \textit{completely unsupervised} approach for video denoising. Here we modify (\ref{loss-re})
	by replacing ${I}_{t}$ with ${\Psi}_{\theta}({\hat{I}}_{t})$ as follows: 
	\begin{equation}\label{loss-re-unsuper}
	{\cal L}_{st} = \Sigma_{i} \| {\Phi}_{ \gamma}\left( \{{\Psi}_{\theta} ( {\hat{I}}_{t+k}(x) ) \}_{k=-K}^{K} \right) - {\Psi}_{\theta}({\hat{I}}_{t}) \|_2. 
	\end{equation}
	This avoids the use of ground truth as clean signal.
	
	Meanwhile, during training we enlarge the decay coefficient $\alpha$ to ensure that there is enough iterations
	for the first stage to learn to map a better ${\Psi}_{\theta}({\hat{I}}_{t})$, which would provide 
	cleaner signal for the second stage.

	\section{Experimental Results}
	\begin{table}[!t]
		\begin{center}
			\begin{tabular}{l c c c c}
				\toprule[1pt]
				\multirow{2}{*}{{Method}}    & \multicolumn{2}{c}{{Vimeo-Gauss25}}  & \multicolumn{2}{c}{{Vimeo-Mixed}}      \\
				\cline{2-5}
				& PSNR                & SSIM                       & PSNR                    &  SSIM                 \\
				\hline\hline
				TOFlow                             & 33.14               & 0.9119                    &  33.40                 &  0.9126    \\
				FITVNet (base)                         & 33.17               & 0.89                        &  32.62                 &  0.88                  \\
				FITVNet (+jsc)                         &  \textbf{34.33}    & \textbf{0.9208}      & \textbf{ 33.93}    & \textbf{0.9168}  \\
				\toprule[1pt]
			\end{tabular}
		\end{center}
		\caption{Quantitative results. Compared with TOFlow, our FITVNet (+jsc) model achieves consistently better results.}
		\label{table-comparison-with-toflow}
	\end{table}
	
	We utilize the large-scale, high-quality Vimeo90K dataset, which is built along with TOFlow~\cite{xue2019video}, for benchmarking our video denoising model, 
	It consists of 89,800 videos of size 256x448, which covers various real-world 
	scenes and actions. We randomly choose 700 videos from whole test dataset, denoted as Vimeo700, 
	for the following testing experiments as the original test dataset is too large.
	
	\subsection{Setup}
	
	In all experiments, we use two kinds of noises: AWGN of $\sigma \in [5, 80]$ and mixed noise of 
	Gaussian noise with standard deviation of 25.5 and 10\% salt-and-pepper noise (Vimeo-Mixed), to train models for comparison with previous methods,
	including flow-based networks and end-to-end training networks. 
	We also train two models.
	The first is trained with the Vimeo-Mixed noise for a fair comparison with TOFlow~\cite{xue2019video} and for demonstrating that our model is capable to deal with such kind of noise. The other is
	trained with the AWGN noise for a fair comparison with FastDVDNet~\cite{tassano2019fastdvdnet} and for demonstrating the 
	superiority of our proposed prior image denoising stage.
	
	\begin{figure}[!t]
		\begin{center}
			\includegraphics[width=\columnwidth]{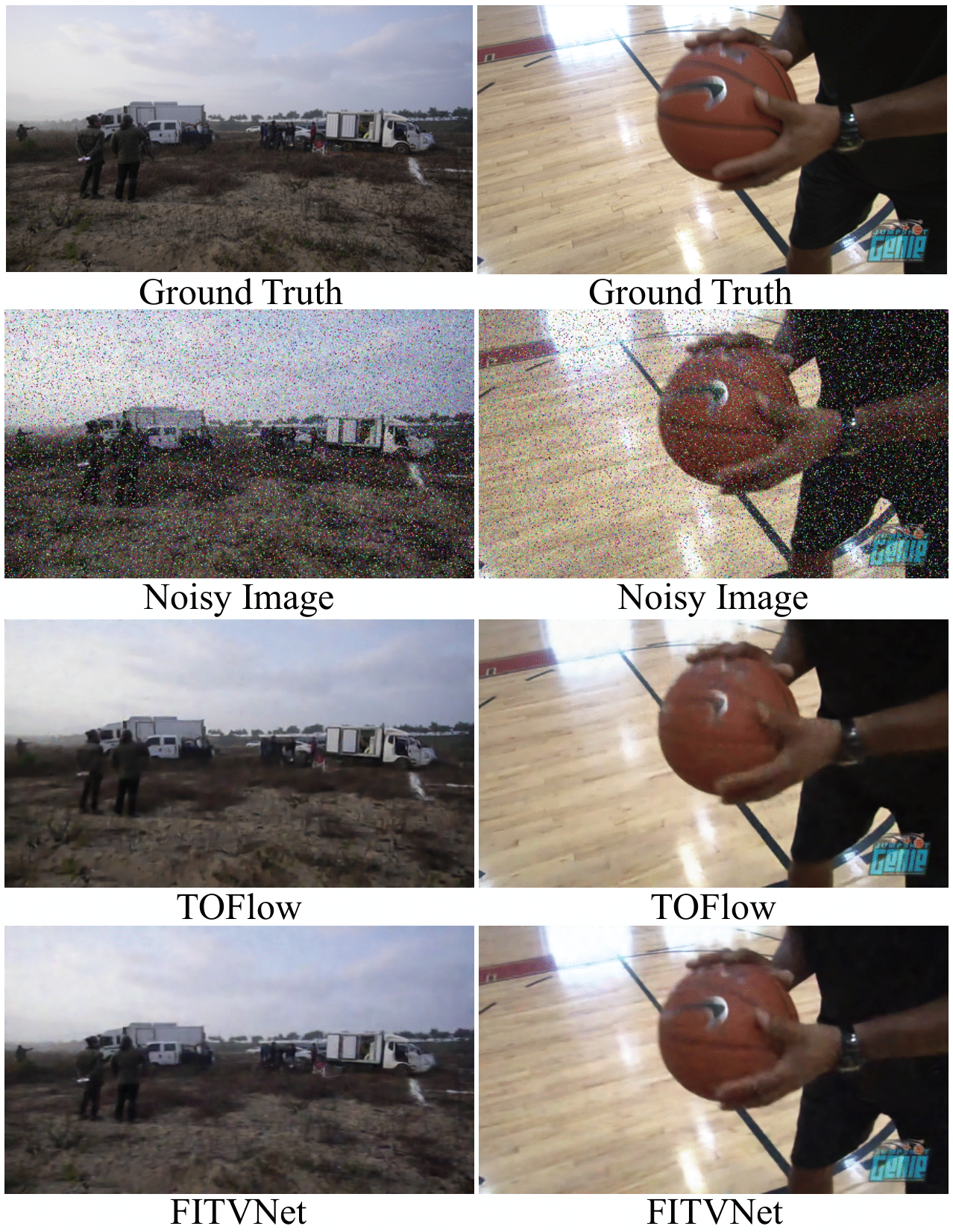}
		\end{center}
		\caption{The denoised results of our model and TOFlow on Vimeo700 with mixed noise. Details 
			are over-smooth in the TOFlow results while our model could recover these more clearly.}
		\label{fig-results-comparison-with-toflow}
	\end{figure}
	
	In our experiments, 
	we randomly choose $2K+1$ frames 
	from the original video if the total number of frames of that video exceeds $2K+1$.
	We do not employ any pretraining procedure used in TOFlow\cite{xue2019video}. 
	We use the commonly used metrics of Peak Signal-to-Noise Ratio (PSNR) and Structural SIMilariry index (SSIM) to characterize our denoising performance. 
	
	The whole network with two models is implemented in  
	PyTorch~\cite{paszke2017automatic}  with a mini-batch of size 16. 
	We optimize the final loss function $\alpha{\cal L}_{sd} + {\cal L}_{st}$ via 
	ADAM~\cite{kingma2014adam} optimizer with default hyperparameters.
	In addition, the total number of epochs is by default set to 40 (unless further explained) and the learning rate is set to be 0.0001.
	
	\subsection{Comparison with flow-based network}
	In the following, we compare our model with TOFlow, which 
	is a state-of-the-art flow-based video denoising network. Table~\ref{table-comparison-with-toflow} 
	quantitatively reports 
	experimental results, which demonstrate the capability of our model to 
	deal with mixed noise in addition to commonly used Gaussian noise.
	The model has been trained on Vimeo-Mixed for 40 epochs with hyperparameters mentioned before.
	Compared with TOFlow, our FITVNet (base) model achieves comparable performance in terms of both PSNR and SSIM and our FITVNet (+jsc) model achieves consistently better results.
	
	Also our two-stage video denoising method alleviates the over-smooth phenomenon happened in TOFlow as we can observe in Figure \ref{fig-results-comparison-with-toflow}.
	For example, in the left column of Figure \ref{fig-results-comparison-with-toflow}, clothes that these two people wear are green, which are similar to the background around them, and the denoised results by TOFlow could not recover such details in this area. Similar phenomenon has also happened even worse in the
	right column of results, where the basketball holder moves fast, 
	and the logo of that basketball is too obscure in the denoised video of TOFlow. In contrast, such details are obviously clearer in our results. 
	
	\begin{figure}[!t]
		\begin{center}
			\includegraphics[width=\columnwidth]{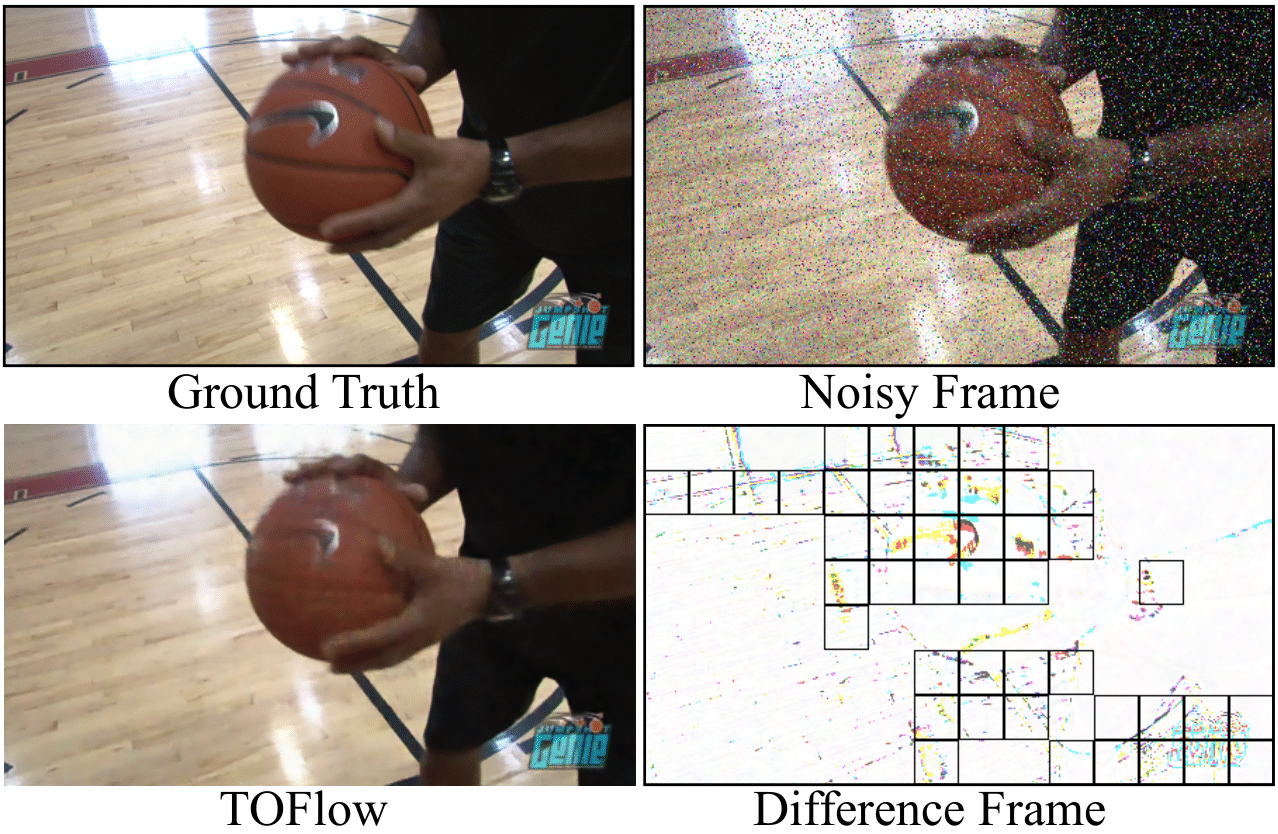}
		\end{center}
		\caption{Here we show the difference image between a denoised frame of TOFlow and its corresponding clean frame and use black boxes to bound patches whose std is higher than the average std 0.0222. 
			Resulted bounding boxes successfully bound the boundaries of the person, basketball, and the logo on the basketball.}
		\label{fig-ad}
	\end{figure}
	
	\begin{table}
		\begin{center}
			\begin{tabular}{l c  c  c}
				\toprule[1pt]
				Method                             &     ${AD}_{avg}$  &  ${AD}_{max}$       &  \#$AD$\\
				\hline\hline
				TOFlow                             &     0.0263          &   0.0617                &  304\\
				FITVNet (+jsc)                    &\textbf{0.0260}  &   \textbf{0.0585}    &  \textbf{298}\\
				\toprule[1pt]
			\end{tabular}
		\end{center}
		\caption{Quantitative characterization of over-smooth phenomenon. Compared with TOFlow, our FITVNet (+jsc) model is slightly advantageous.}
		\label{ad-comparison-with-toflow}
	\end{table}

	\begin{table}
		\begin{center}
			\begin{tabular}{l c }
				\toprule[1pt]
				Method                             &    Time (s)                         \\
				\hline\hline
				TOFlow                             &     0.2460                            \\
				FITVNet (+jsc)                         &   \textbf{0.0466}          \\
				\toprule[1pt]
			\end{tabular}
		\end{center}
		\caption{Comparison of running time to denoise an RGB frame of 
			resolution 256x448. Compared with TOFlow,  our FITVNet (+jsc) model is more than 5 times faster.}
		\label{time-comparison-with-toflow}
	\end{table}

	\begin{table*}[!t]
		\begin{center}
			\begin{tabular}{l c c c c c c c c c c}
				\toprule[1pt]
				\multirow{2}{*}{{Method}}    & \multicolumn{2}{c}{{Vimeo-Gauss15}}  &\multicolumn{2}{c}{{Vimeo-Gauss25}}   &\multicolumn{2}{c}{{Vimeo-Gauss35}} & \multicolumn{2}{c}{{Vimeo-Gauss45}} & \multicolumn{2}{c}{{Vimeo-Gauss55}}\\ 
				\cline{2-11}
				& PSNR                & SSIM                       & PSNR            & SSIM &                                        PSNR                & SSIM                       & PSNR            & SSIM
				&                                        PSNR               & SSIM\\
				\hline\hline
				TOFlow                          & 31.64          & 0.8719                    & 33.14          & 0.9195
				&                                         32.70               & 0.8849                    & 30.70         &   0.8352
				&                                         28.59               & 0.7732\\
				FastDVDNet                       & 33.85            & 0.9114                    & 32.13         & 0.8754                             &                                            30.77            & 0.8387                    & 29.63        &  0.7990
				&                                            28.70            & 0.7589\\
				FITVNet (base)                         & 30.73            & 0.8664                    & 35.21          & 0.9248                   &                                         33.78               & \textbf{0.9118}        & 32.64          & \textbf{0.8935}
				&                                         \textbf{31.73}    & \textbf{0.8760}\\
				FITVNet (+jsn)                         &36.76             & 0.9475                  & 34.80         & 0.9270                           &                                          33.41              &  0.9078                   & 32.34         & 0.8896
				&                                          31.44               & 0.8723\\
				FITVNet (+jsc)                        &\textbf{37.70}  &  \textbf{0.9552}    &\textbf{35.45} & \textbf{0.9326}                &                                       \textbf{33.89}     & \textbf{0.9118}     &\textbf{32.68}  & 0.8926		 	    &                                          31.72                & 0.8744\\
				\hline\hline
				FITVNet (unsupervised)             & 34.23       & 0.8944                &   32.79       & 0.8743
				&                                           31.72               & 0.8557                 &  30.83         & 0.8377          
				&                                           30.02               & 0.8202\\       
				\toprule[1pt]
			\end{tabular}
		\end{center}
		\caption{Quantitative results. Compared with TOFlow and FastDVDNet, our baseline FITVNet  model is consistently better except that our base model fails with the 'Gauss15' setting, while our other FITVNet models are robust when supervised with ${\cal L}_{pd}$.}
		\label{table-comparison-with-fastdvd}
	\end{table*}
	
	To quantitatively characterize such
	a problem existing in TOFlow, we propose to compute Average Deviation (AD). Based on the difference 
	image ${I}_{diff}$ between each predicted image $\hat{I}$ and the corresponding clean image $I$, we first divide the image ${I}_{diff}$ into $K$ nonoverlapping patches of size $32*32$ and calculate the standard deviation $\rho(k)$ for each patch $k$. Then, we select those patches whose $\rho(k)$ is larger than a threshold $\bar \rho$ we set as the average of $\{\rho(k);k=1:K\}$, denoted by $\bar \rho$. Finally, we compute the average of the selected $\rho(k)$ as the AD index.
	\begin{equation}
	AD = \frac{\sum_{k:\rho(k)> {\bar \rho} } \rho(k)} {\sum_{k:\rho(k)> {\bar \rho}} 1}.
	\end{equation}
	
	
	When the over-smooth phenomenon happens along the boundaries of fast-moving objects, the difference image has structural residuals, which contribute to high values of $\rho(k)$, such that the phenomenon are likely captured by the proposed AD index.
	In Figure~\ref{fig-ad}, we also visually 
	show this. 
	Specifically, we use black boxes to
	bound those selected patches, 
	Obviously, we successfully localize the  boundaries of the fast moving objects such as basketball, logo and the person, which coincides with our visual perception.
	
	In experiments, we compute the average of $AD$ (${AD}_{avg}$) and the max of $AD$ (${AD}_{max}$) of all 
	test images on Vimeo700. We also tally the total number of test images (\#$AD$) whose $AD$ is larger than ${AD}_{avg}$. The above three statistics are shown in Table~\ref{ad-comparison-with-toflow}.
	Although ${AD}_{avg}$ is similar for these two models, TOFlow yields higher values of ${AD}_{avg}$, ${AD}_{max}$ and \#$AD$, which indicates that TOFlow likely suffers from the over-smoothing problem than FITVNet. This concides with Figure~\ref{fig-results-comparison-with-toflow} too. 
	
	\begin{figure}[t]
		\begin{center}
			\includegraphics[width=\columnwidth]{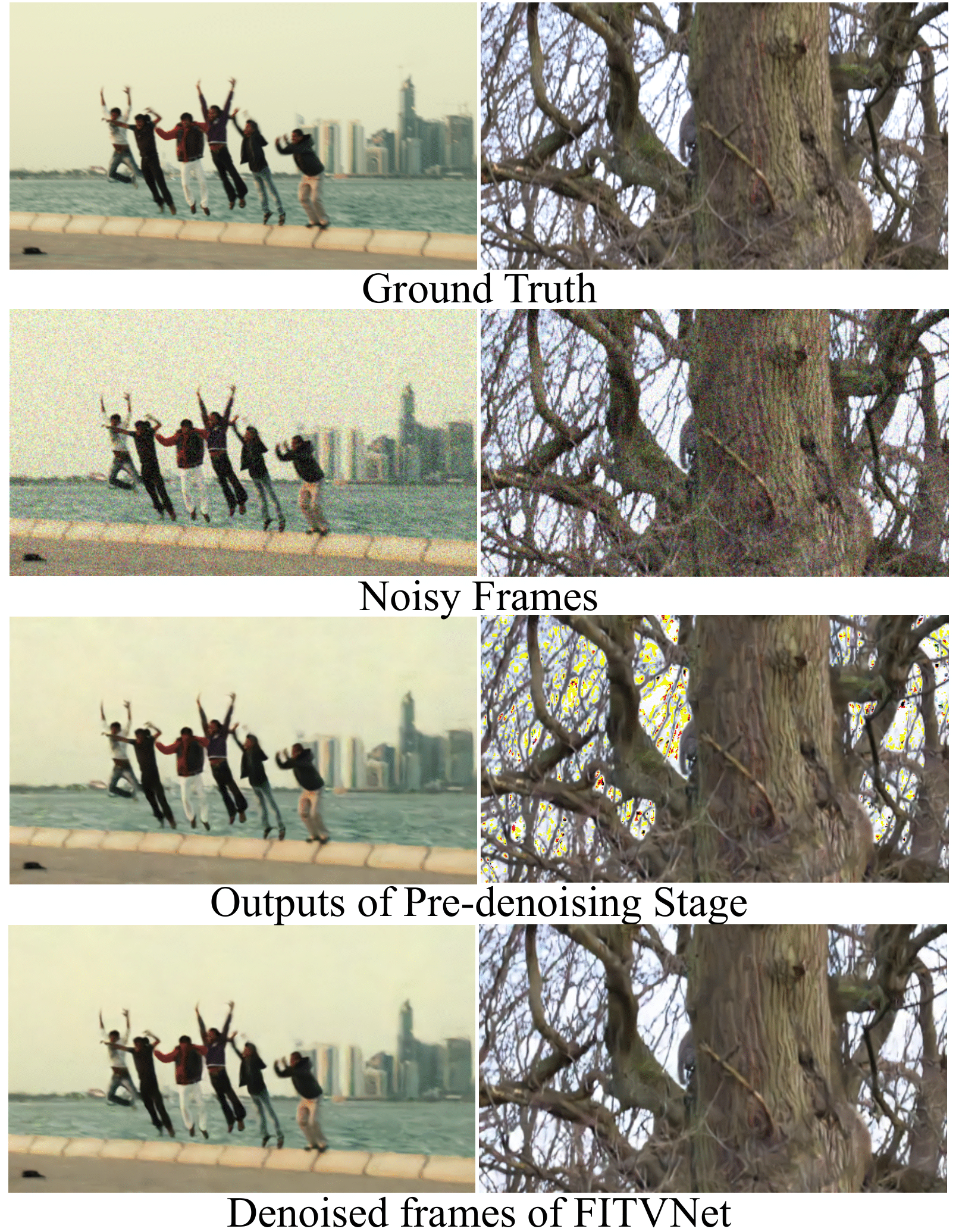}
		\end{center}
		\caption{Output images of the prior image denoising module and the final denoised frames. 
			The output images are clear, and spatial noise is largely reduced, fostering the following video denoising module to learn better results.}
		\label{fig-features}
	\end{figure}
	
	\textbf{Running time.} TOFlow improves traditional 
	flow-based networks to utilize a convolutional network to predict
	optical flow of neighbouring frames. However, the warping
	operation also takes more computation in the whole process. In
	contrast, our FITVNet method achieves fast inference with the help 
	of direct network implementation. As shown 
	in Table~\ref{time-comparison-with-toflow}, it takes only 46.6ms to denoise
	a 256x448 RGB frame, while TOFlow needs 246ms. The experiments are
	all run on the same GPU of NVIDIA 2080ti card. 
	
	
	\subsection{Comparison with end-to-end methods}
	Then we compare our model with FastDVDnet~\cite{tassano2019fastdvdnet}, which is a 
	spatiotemporal video denoising algorithm, based on Vimeo700 with AWGN of different levels.
	The models are all trained on Vimeo90K, and results are shown in Table~\ref{table-comparison-with-fastdvd}.
	It is obvious that our three variants perform much better than FastDVDNet, even though
	our base model is only supervised with ${\cal L}_{st}$. Besides, we also test TOFlow 
	on Vimeo700 with Gaussian noise with the TOFlow's open checkpoint model since we have no access to TOFlow's training
	process. The TOFlow results are robust, but not as good as the performance on Vimeo-Mixed.
	Although our FITVNet (base) model performs not very well with `Gauss15', the other two variants, 
	FITVNet (+jsc) and FITVNet (+jsn), are stable across all noise levels with the help of the ${\cal L}_{pd}$ loss.
	
	\begin{figure*}
		\begin{center}
			\includegraphics[width=\textwidth]{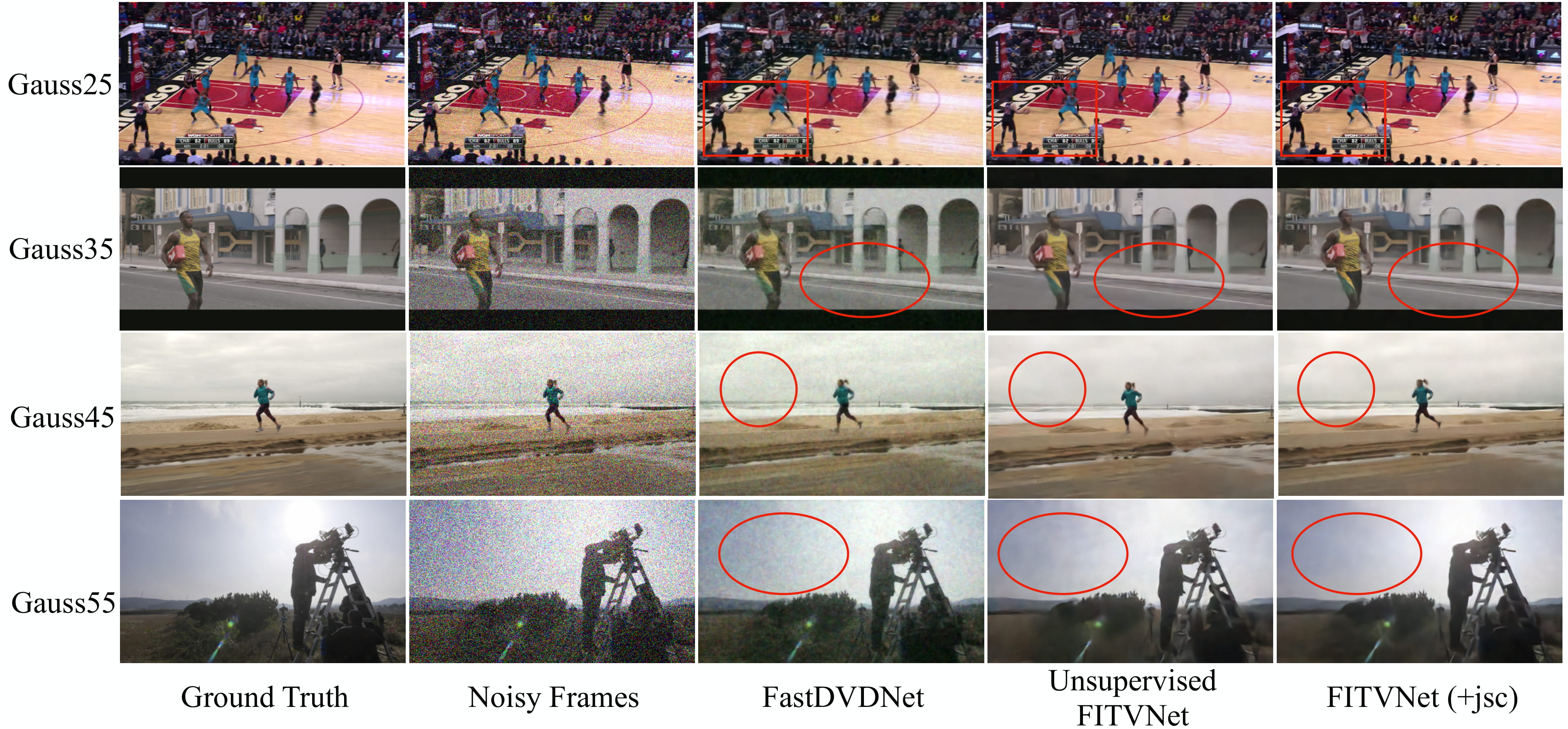}
		\end{center}
		\caption{We qualitatively show the results of FastDVDNet, our best model FITVNet (+jsc), and our unsupervised
			FITVNet. FITVNet (+jsc) performs the best on all these cases, no matter with fast-moving objects or with 
			highly noisy backgrounds. Besides, unsupervised FITVNet performs comparablely with FastDVDNet. FastDVDNet
			could recover more details while unsupervised FITVNet is capable of removing more spatial noise with help of prior image denoising 
			module.}
		\label{fig-fast}
	\end{figure*}
	
	To further demonstrate the effect of prior image denoising module, we visualize the outputs of this module in Figure~\ref{fig-features}. From the left subfigure, we could
	observe that the output image is already clean so that the following spatiotemporal module would just learn to model object motion this time. On the other hand, in the case shown in 
	the right subfigure, the prior image denoising module does not remove all spatial noise, and the following
	spatiotemporal module would learn to model both spatial noise and temporal deformation this time.
	These two cases characterize how the first and the second modules cooperate with each other to accomplish video denoising tasks with state-of-the-art results.
	
	Furthermore, we qualitatively show in Figure~\ref{fig-fast} the denoised frames of our models and the FastDVDNet
	at different noise levels. It can be easily perceived that 
	the FITVNet results are visually more pleasing than the FastDVDNet results. For the background regions where we bound with red boxes and circles, our results evidently possess higher visual quality. Furthermore, along the boundaries of moving objects as in the  `Gauss25' setting, the results appear sharper.  
	
	\subsection{Unsupervised experiment} We also test our unsupervised FITVNet as 
	introduced in Section \ref{unsupervised_FITVNet} and the results are also shown in Table~\ref{table-comparison-with-fastdvd}. 
	Although its performance is not as good as FITVNet with clean frames as supervision, it achieves better performance than FastDVDNet across the board in terms of PSNR and comparable performance to TOFlow on Vimeo700. Note that both FastDVDNet and TOFlow are supervised with clean frames. The running time of unsupervised FITVNet is the same as that of other FITVNet models.
	
	Besides, we also qualitatively show our denoised frames in Figure~\ref{fig-fast}.
	It is evident that our denoised frames exhibit less spatial noise when compared with FastDVDNet, which emphasizes
	the importance of prior image denoising module. The success of our completely unsupervised denoising 
	model also demonstrates the power of our simple yet effective idea for such a two-stage video denoising framework.
	\section{Conclusion}
	In this work, we propose a two-stage approach called FITVNet for video denoising task, which first reduces the noise frame by frame and then applies spatiotemporal denoising. This aims to address the challenge that the boundaries of fast moving objects in the denoised results obtained by previous models, which either estimate object motion via optical flow or directly apply spatiotemporal denoising, tend to be blurry.
	In addition, our model is supervised and is learned in an end-to-end manner; but it also renders the possibility of a purely unsupervised learning approach. Our extensive evaluations based on the Vimeo90K dataset demonstrate that the proposed FITVNet recovers object boundaries more
	clearly and achieve state-of-the-art denoising results in terms of quantitative measures such as PSNR and SSIM while running over 5 times faster than TOFlow. Furthermore, 
	the purely unsupervised FITVNet, implemented by replacing 
	supervision signal of the second module with the output of the first module, achieves comparable denoising performance to FastDVDNet.
	
	The failure of our base model with the `Gauss15' setting in Table \ref{table-comparison-with-fastdvd} suggests that implicit object motion estimation with a direct spatiotemporal networks is not as stable as flow-based model. In future, we plan to explore the possibility of integrating flow calculation into our FITVNet for improved robustness and performance. We also plan to explore more along unsupervised video denoising.
	\clearpage
	
	
	\clearpage
		{\small
		\bibliographystyle{ieee\_fullname}
		\bibliography{egbib-ori} }
	
	\clearpage	
	\section{Appendix}
	
	To make the experimental setup more clearly, we would display detailed network architectures 
	of two modules in FITVNet. In the meanwhile, we would exhibit more denoised examples.
	
	\begin{table}[b]
		\begin{center}
			\begin{tabular}{l|l|l}
				\toprule[1pt]
				Name                              &  ${N}_{out}$          &  Description                      \\
				\hline
				Input                               &  12                      & Three frames with noise map      \\
				Enc\_Conv1a                 &  3x30                  &  Convolution 3x3                  \\
				Enc\_Conv1b                 &  32                      &   Convolution 3x3                 \\
				Enc\_Conv1c                 &  64                      &  Convolution 3x3   Stride 2    \\
				Enc\_Conv2a                 &  64                      &  Convolution 3x3                  \\
				Enc\_Conv2b                 &  64                      &  Convolution 3x3                  \\
				Enc\_Conv2c                  &  128                    &  Convolution 3x3  Stride 2     \\
				Enc\_Conv3a                  &  128                    &  Convolution 3x3                  \\
				Enc\_Conv3b                  &  128                    &  Convolution 3x3                  \\
				\hline
				Dec\_Conv3a                  &  128                    &  Convolution 3x3                  \\
				Dec\_Conv3b                  &  128                    &  Convolution 3x3                  \\
				Dec\_Conv3c                  &  256                    &  Convolution 3x3                  \\
				Dec\_Upsample3                &  64                      &  PixelShuffle                        \\
				Dec\_Plus3                        &  64                      &  Plus Output of Enc\_Conv2b        \\
				Dec\_Conv2a                  &  64                      &  Convolution 3x3                  \\
				Dec\_Conv2b                  &  64                      &  Convolution 3x3                  \\
				Dec\_Conv2c                  &  128                    &  Convolution 3x3                  \\
				Dec\_Upsample2                &  32                     &   PixelShuffle                        \\
				Dec\_Plus2                        &  32                      &  Plus Output of Enc\_Conv1b          \\
				Dec\_Conv1a                  &  32                      &  Convolution 3x3                  \\
				Dec\_Conv1b                  &  3                        &  Convolution 3x3                  \\
				Dec\_Plus1                        &  3                        &  Plus Reference frame            \\
				\toprule[1pt]
			\end{tabular}
		\end{center}
		\caption{Network Architecture of spatiotemporal video denoising module.}
		\label{network2}
	\end{table}

	\subsection{Network Architectures}
	The proposed FITVNet consists of three modified U-Net~\cite{ronneberger2015u}, of which one is
	used for the spatial image denoising module and presented in Table~\ref{network1}, while the other two 
	are for spatiotemporal video denoising module and displayed in Table~\ref{network2}. 
	Note that ${N}_{out}$ denotes the number of output channels and each convolution layer is followed by a ReLU activation layer (unless further explained).
	
	\subsection{Comparison with TOFlow and FastDVDNet via Sobel operator}
\begin{figure}
		\centering
		\begin{minipage}[t]{0.235\textwidth}
			\centering
			\includegraphics[width=1\textwidth]{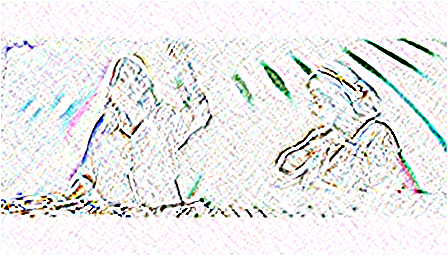}
			FastDVDNet~\cite{tassano2019fastdvdnet}
		\end{minipage}
		\centering
		\begin{minipage}[t]{0.235\textwidth}
			\centering
			\includegraphics[width=1\textwidth]{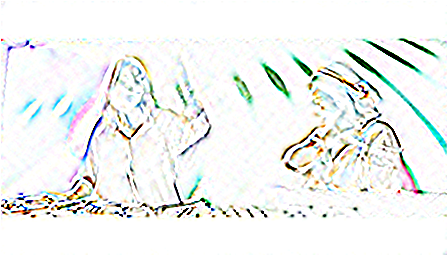}
			TOFlow~\cite{xue2019video}
		\end{minipage}
		\label{sobel}
	\end{figure}

	In Section 4.2, We have showed the performance of object boundaries recovery of FITVNet, TOFlow~\cite{xue2019video}, and FastDVDNet~\cite{tassano2019fastdvdnet} with our proposed AD index. Next, we further exhibit processed results with Sobel operator in Figure~\ref{sobel}. The operator is a traditional boundary detection method, which gives the direction of the largest possible increase from light to dark and the rate of change in that direction according to calculations of the gradient of the image intensity at each point.

	As results shown in Figure~\ref{sobel}, FastDVDNet is the worst one since Sobel is not noise sensitive, i.e., the denoised results of FastDVDNet are still with a higher noise level. Although TOFlow and FITVNet both show high performance on denoising these videos, FITVNet deals with object boundaries of fast-moving objects, i.e. the woman in the shown video, better than TOFlow.
	
	\begin{figure}
		\centering
		\begin{minipage}[t]{0.235\textwidth}
			\centering
			\includegraphics[width=1\textwidth]{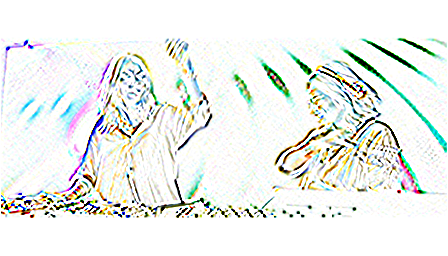}
			Ground Truth
		\end{minipage}
		\centering
		\begin{minipage}[t]{0.235\textwidth}
			\centering
			\includegraphics[width=1\textwidth]{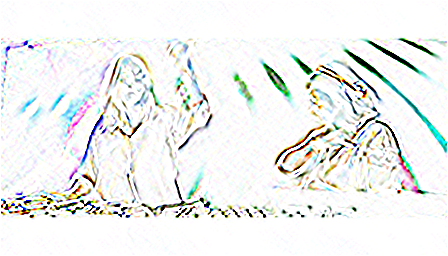}
			FITVNet
		\end{minipage}
	\end{figure}
	
\begin{table}[b]
	\begin{center}
		\begin{tabular}{l | l | l}
			\toprule[1pt]
			Name                           &   ${N}_{out}$               & Discription  \\
			\hline
			input                            &   3                            & One frame\\
			Enc\_Conv0                  &   48                           &Convolution 3x3\\
			Enc\_Conv1a                &   48                           &Convolution 3x3\\
			Enc\_Conv1b                &   48                           &Convolution 3x3 Stride 2\\
			Enc\_Conv2a                &   48                           &Convolution 3x3\\ 
			Enc\_Conv2b                &   48                           &Convolution 3x3\\ 
			Enc\_Conv3a                &   48                           &Convolution 3x3 Stride 2\\ 
			Enc\_Conv3b                &   48                           &Convolution 3x3\\ 
			Enc\_Conv4a                &   48                           &Convolution 3x3\\ 
			Enc\_Conv4b                &   48                           &Convolution 3x3 Stride 2\\ 
			Enc\_Conv5a                &   48                           &Convolution 3x3\\ 
			Enc\_Conv5b                &   48                           &Convolution 3x3\\ 
			Enc\_Conv6                     &   48                           &Convolution 3x3 Stride 2\\ 
			\hline
			Dec\_Upsample5               &  48                           &Upsample 2x2\\ 
			Dec\_Concat5                   &  96                           &Concatenate Output of Enc\_Conv4b\\ 
			Dec\_Conv5a                 &  96                           &Convolution 3x3\\ 
			Dec\_Conv5b                 &  96                           &Convolution 3x3\\ 
			Dec\_Upsample4               &  96                           &Upsample 2x2\\  
			Dec\_Concat4                   &  144                         &Concatenate Output of Enc\_Conv3b\\ 
			Dec\_Conv4a                 &  96                           &Convolution 3x3\\ 
			Dec\_Conv4b                 &  96                           &Convolution 3x3\\  
			Dec\_Upsample3               &  96                           &Upsample 2x2\\  
			Dec\_Concat3                   &  144                         &Concatenate Output of Enc\_Conv2b\\ 
			Dec\_Conv3a                 &  96                           &Convolution 3x3\\
			Dec\_Conv3b                 &  96                           &Convolution 3x3\\ 
			Dec\_Upsample2               &  96                           &Upsample 2x2\\ 
			Dec\_Concat2                   &  144                         &Concatenate Output of Enc\_Conv1b\\ 
			Dec\_Conv2a                 &  96                           &Convolution 3x3\\ 
			Dec\_Conv2b                 &  96                           &Convolution 3x3\\ 
			Dec\_Upsample1               &  96                           &Upsample 2x2\\ 
			Dec\_Concat1                   &  96+3                       &Concatenate Input\\ 
			Dec\_Conv11                 &  64                           &Convolution 3x3\\ 
			Dec\_Conv11                 &  32                           &Convolution 3x3\\ 
			Dec\_Conv0                      &  3                           &Convolution 3x3 LeakyReLU($\alpha$=0.1)\\ 
			\toprule[1pt]
		\end{tabular}
	\end{center}
	\caption{Network Architecture of spatial image denoising module.}
	\label{network1}
\end{table}

	\clearpage
	
\end{document}